\documentclass[letterpaper]{article} 
\usepackage{aaai2026}  
\usepackage{times}  
\usepackage{helvet}  
\usepackage{courier}  
\usepackage[hyphens]{url}  
\usepackage{graphicx} 
\usepackage{natbib}  
\usepackage{caption} 
\usepackage{algorithm}
\usepackage{algorithmic}

\usepackage{newfloat}
\usepackage{listings}
\DeclareCaptionStyle{ruled}{labelfont=normalfont,labelsep=colon,strut=off} 
\floatstyle{ruled}
\newfloat{listing}{tb}{lst}{}
\floatname{listing}{Listing}
\usepackage{listings}
\lstdefinestyle{leanstyle}{
  language=lean,
  basicstyle=\ttfamily\small,
  keywordstyle=\bfseries\color{blue},
  commentstyle=\itshape\color{gray},
  breaklines=true,
  columns=fullflexible,
  keepspaces=true,
  showstringspaces=false,
  frame=single,
  framerule=0.8pt,
  backgroundcolor=\color{gray!5},
  captionpos=b
}
\usepackage{tcolorbox}
\tcbuselibrary{breakable}
\lstdefinelanguage{lean}{
  keywords={
    theorem, lemma, def, inductive, structure, namespace, variable, variables,
    section, end, constant, constants, begin, end, assume, example, instance,
    open, import, if, then, else, forall, exists, fun, match, with, do,
    by, have, show, from, exact, let, in, return, Type, Prop, sort
  },
  sensitive=true,
  comment=[l]--,
  morecomment=[s]{/-}{-/},
  morestring=[b]",
}
\usepackage{xurl}
\usepackage{booktabs}
\usepackage{amsmath}
\usepackage{amssymb}
\usepackage{mathtools}
\usepackage{amsthm}
\usepackage{microtype}
\usepackage{graphicx}
\usepackage{subfigure}
\usepackage{booktabs}
\title{LeanTutor: Towards a Verified AI Mathematical Proof Tutor}
\author {
    Manooshree Patel,
    Rayna Bhattacharyya \equalcontrib, 
    Thomas Lu \equalcontrib, 
    Arnav Mehta \equalcontrib, 
    Niels Voss \equalcontrib, 
    Narges Norouzi, 
    Gireeja Ranade
}
\affiliations {
    University of California, Berkeley \\
    \{manooshreepatel, rayna\_b, thomaslu, arnavmehta, niels\_voss, norouzi, gireeja\}@berkeley.edu
}

\begin{document}

\maketitle

\begin{abstract}
This paper considers the development of an AI-based provably-correct mathematical proof tutor.
While Large Language Models (LLMs) allow seamless communication in natural language, they are error prone. 
%
 Theorem provers such as Lean allow for provable-correctness, but these are hard for students to learn. We present a proof-of-concept system (LeanTutor) by combining the complementary strengths of LLMs and theorem provers. 
%
LeanTutor is composed of three modules: (i) an autoformalizer/proof-checker, (ii) a next-step generator, and (iii) a natural language feedback generator. To evaluate the system, we introduce PeanoBench, a dataset of 371 Peano Arithmetic proofs in human-written natural language and formal language, derived from the Natural Numbers Game. 

\end{abstract}


\section{Introduction}
\label{sec:introduction}
\begin{figure*}[btp]
\centering
\includegraphics[width=01.0\linewidth]{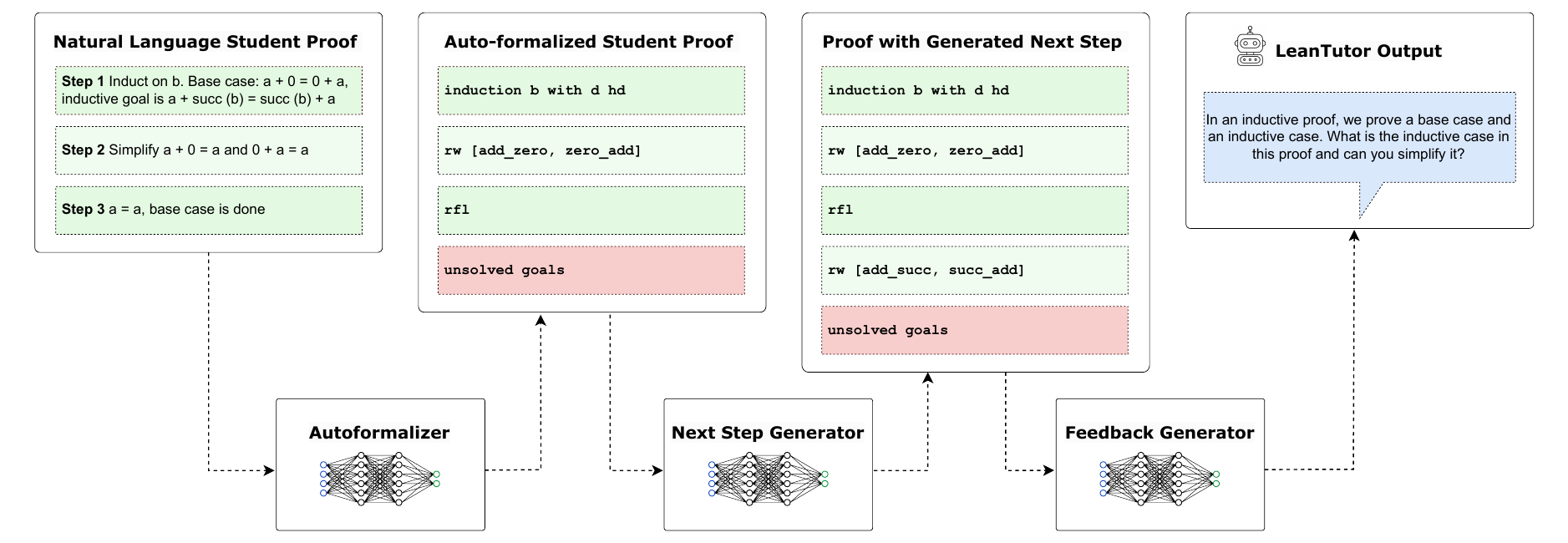}
\caption{LeanTutor is comprised of three modules: an autoformalizer that automatically formalizes an NL student proof into Lean step-by-step; a next step generator that generates a next feasible tactic for the student proof; and a natural language feedback generator that generates guiding feedback to help the student progress towards a correct proof.}
\label{fig:systemdiagram}
\end{figure*}

College students use LLMs such as ChatGPT and Claude to start projects, create practice questions, and generate solutions to academic assignments \citep{openai2025college, anthropic2025education}. 
However, indiscriminate LLM usage can be detrimental to student learning \citep{goetze2025students}, because these systems are not designed from a pedagogical perspective. Specifically, (1) most models are designed to be maximally ``helpful'' to a user \citep{askell2021general}, and often directly give away the answer,
instead of helping a student come up with it on their own~\cite{sonkar2024pedagogical},  
(2) even state-of-the-art models are prone to hallucinations and generate convincing wrong answers \citep{balunovic2025mathconstruct, maurya2024unifying, gupta2025beyond}, (3) models struggle to identify mistakes in reasoning~\citep{tyen2024llms, miller2024llm}, and (4) even if models can produce the correct answer, they cannot necessarily produce correct reasoning to guide the student \citep{gupta2025beyond}. As a result, faculty are increasingly concerned about the negative impacts of LLMs on student learning, particularly in courses where ``critical thinking'' is one of the key learning goals, for example mathematical proof courses. The recent CRA Practioner-to-Professor survey reinforced the importance of math for developing critical thinking~\cite{cra2025p2p}.


There is a desire among educators to find a happy-medium solution --- where a student can get the benefits of the instantaneous, private feedback that LLMs offer, while mitigating the negative effects of LLMs.
Math educators have been developing intelligent tutoring systems (ITS) (e.g.~\citep{bundy2000intelligent, lodder2021generation, barnes2008toward}) and 
theorem prover-based tutors (e.g.~\citep{sieg2007apros, wemmenhove2022waterproof}) to help students do math proofs for decades.
Unlike LLMs these systems are 
definitively correct, but they can be tedious to create \citep{dermeval2018authoring}. Others require an understanding of challenging formal language syntax \citep{thoma2022learning}, or constrict student writing by only accepting input in application-specific controlled natural language \citep{wemmenhove2022waterproof}. As a result of these challenges, we do not see widespread adoption of such tutoring systems (unlike autograders for programming~\citep{denero2014teaching, hecht2023distributing, mitra2023studying}). 

We propose LeanTutor, a math-proof tutoring system that combines the strengths of LLMs and theorem provers. This system interacts with students in natural language (NL), while using the Lean theorem prover \citep{moura2021lean} to evaluate proof correctness and generate correct next steps on the backend. 
Specifically, the system can: 
\begin{itemize}
    \item Accept complete/partial/correct/incorrect student  proofs
    \item Verify if the student work is correct or incorrect
    \item Identify the student error, if applicable, and provide guidance towards a correct proof, without giving away the complete answer.
\end{itemize}

This paper presents a first design of LeanTutor, working off of a small self-contained dataset, as in~\citep{murphy2024autoformalizing, cunningham2023towards}. 
This system requires a seamless back-and-forth between the LLM and theorem prover by means of faithful auto-formalization and auto-informalization. However, compared to the standard approach in AI for Math work, the tutoring setting offers a new frame for the problems of auto-formalization and next-step generation, as below.

\begin{itemize}
\item Autoformalization for tutoring systems must focus on \textit{faithful autoformalization} of natural language statements (i.e. preserving the semantic meaning as in~\citep{murphy2024autoformalizing}). Fundamental differences in how informal and formal math is written make this difficult. First, formal proofs are much more fine-grained than informal proofs --- proof steps that would be considered rigorous by humans (e.g. arguments that are true by symmetry) can require long-winded proofs in formal languages. 
%
Second, informal proofs commonly use forward-reasoning where are formal proofs can be often constructed using backwards-reasoning \citep{shi2025qed}. This difference of style can create significant challenges. 

\item Evaluating faithful autoformalization is particularly challenging, as syntactic correctness (i.e. successful compilation) is insufficient \citep{liurethinking}. 

\item Autoformalization for a tutoring system must be able to handle not only complete and correct but incomplete and incorrect proofs. Student natural language further have lots of variation and may not have the polish of professional mathematical writing.  Previous work on autoformalization has focused on theorem statements and correct whole proofs~\citep{yang2024formal}. 

\item The tutoring application may assume knowledge of a complete proof of the theorem under consideration, creating an easier version of the faithful autoformalization problem. Students are working on problems where the instructor knows a solution, unlike mathematicians trying to solve novel problems. We see that this improves  autoformalization performance, and leaves open future research problems on how to fully exploit a reference solution. 

\item Similarly, this reference proof means that the key challenge in next-step-generation is to identify the specific proof approach taken by the student, which may or may not match that of the reference solution. However, the tutoring application can limit the search for next steps to relatively small theorem libraries, compared to novel proof search using huge libraries such as Mathlib~\citep{mathlib2020}.

\end{itemize}

LeanTutor simplifies some of these challenges through a carefully aligned informal-formal math dataset, discussed in Section \ref{sec:peanobench}, and by proposing a new metric to measure autoformalization accuracy (explained in \ref{sec:ourmetrics}), but by no means solves them. 
The challenges in developing AI-math-tutors are very similar to those in developing general AI-mathematics-assistants~\citep{riehl2025testing} and remain open problems for us to tackle.

\section{Related Work}

\subsection{Autoformalization via Language Models}
A large body of recent work has focused on autoformalizing natural language (NL) theorem statements into formal language (FL), using deep learning methods \citep{ying2024lean, gao2024herald, wu2022autoformalization, azerbayev2023proofnet, lin2025goedel, shao2024deepseekmath}. The more difficult task of autoformalizing whole proofs from NL to FL has been explored in fewer works \citep{jiang2022draft, murphy2024autoformalizing, wang2024theoremllama, huang2024mustard}. State-of-the-art (SOTA) LLMs, without any specific formal language training, have shown strong performance on the task of autoformalization \citep{wu2022autoformalization} and motivate our development of an LLM-agnostic framework for autoformalization. 

Both tutoring (as in LeanTutor) and auto-grading require \textit{faithful autoformalization} \citet{murphy2024autoformalizing}. 
In this paper, we take a similar approach to \citet{kulal2019spoc} method of translating pseudocode to code, line-by-line, in a C++ program generation task. 
We make the reasonable assumptions for the classroom setting that all proofs come from a small dataset and at least one valid proof per theorem is known (in both NL and FL). \citet{murphy2024autoformalizing, cunningham2023towards} successfully formalize proofs in a small dataset where all feasible theorems/tactics are known. 

\subsection{Neural Theorem Proving}
Neural theorem proving reframes theorem proving as a language modeling task \citep{lisurvey}. An abundance of prior work has made progress towards training language models for theorem proving \citep{ wang2024theoremllama, welleck2022naturalprover, azerbayev2023llemma, lin2025goedel, yang2023leandojo}. 
Additionally, prior work has explored theorem proving frameworks with SOTA LLMs \citep{jiang2022draft, huang2024mustard, thakur2023context}.  Our next-step generation approach is largely inspired by the COPRA agent \citep{thakur2023context}. The COPRA agent performs a GPT-4 directed depth-first search over sequences of possible tactics, to complete a formal proof. The agent additionally implements a ``progress check'', which assesses if generated tactics progress the proof. 


\subsection{Automated Feedback Generation for Programming Assignments}
We draw inspiration for LeanTutor's feedback generation module from automated feedback generation in programming classes \citep{d2015can, singh2013automated, suzuki2017exploring}. Since students write their code in a programming environment where compilers enforce formal correctness and autograders ensure that the code passes test cases or return appropriate errors, autonomous tutors can leverage the resulting error messages and metadata to generate high-quality feedback. We build on the five hint types identified by \citet{suzuki2017exploring} (transformation, location, data, behavior, and example) that can be generated via program synthesis to provide students feedback in an introductory coding class. Similarly, theorem provers provide proof state information and specific error messages, which can be used to pinpoint student errors. 


Autoinformalization, translating formal statements into informal ones \citep{lisurvey}, is a parallel task to feedback generation. LLM-based autoinformalization has been explored with success \citep{wu2022autoformalization, huang2024mustard}.

\subsection{Math Proof Tutors}
We identify three categories of existing math proof tutors---intelligent tutoring systems, LLM-based tutors, and theorem prover-based tutors. Researchers have made attempts to develop \citep{autexier2012towards, briggle2008towards} or developed intelligent tutoring systems (ITS) for math proofs \citep{barnes2008toward, lodder2021generation, bundy2000intelligent}. ITS require expert authoring of solutions or feedback, making them difficult to develop and scale \citep{dermeval2018authoring}. LLM-based math tutors have demonstrated benefits such as learning gains \citep{pardos2023learning} and can maintain conversations with no harmful content \citep{levonian2025designing}. However, these LLMs fail as tutors, for the reasons outlined in Section \ref{sec:introduction}. Math educators have used theorem provers, such as Lean, 
to teach proofs \citep{avigad2019learning}. These tools have led to unique benefits in students' learning of proofs \citep{thoma2022learning}, but students struggle to learn the complex syntax required to interact with most \citep{avigad2019learning, buzzard_teaching, karsten2023proofbuddy}. 


A more extensive review of these three categories of tutors can be found in the Appendix. 


\section{PeanoBench Dataset}
\label{sec:peanobench}

To develop LeanTutor, 
we construct the PeanoBench dataset, which contains a total of 371 proofs. Each proof has a human-written natural language proof and a semantically equivalent formal language proof in Lean. PeanoBench is derived from the original 80 Peano Arithmetic (PA) proofs in the Natural Number Game 4 (NNG4)~\citep{nng4} (Apache-2.0 license). We chose these because the formal proofs were already available, making evaluation easier, and the informal versions of PA proofs naturally maintain some similarity in flow and granularity to their formal versions. 

To create the dataset, we began with a subset of 75 of the original NNG4 proofs\footnote{We removed the attempted proof of Fermat's Last Theorem and proofs which contain the \texttt{simp} tactic, as the applicability of the \texttt{simp} tactic is broad and can sometimes easily close complex goal}) and back-translated, following similar approaches for low-resource languages~\cite{ranathunga2023neural,ying2024lean, wang2024theoremllamatransforminggeneralpurposellms, lu2024process}. All NL annotations were written by the first five paper authors. A categorization of selected proofs by topic (proofs in NNG4 are arranged by ``worlds'', which correspond to topics like addition, multiplication, etc.) is in
the Appendix. 

%
Proof annotators followed two rules while annotating. (1) Natural language annotations are free of Lean-specific syntax, premises, or tactics. (2) Natural language annotations are written to function as standalone proofs independent of the Lean code. 
The \textbf{one-to-one correspondence between NL proof steps and individual FL tactics} differentiates PeanoBench from prior datasets for Lean autoformalization that pair whole Lean proofs with their whole NL counterpart \citep{lu2024process, wang2024theoremllama, gao2024herald}. 
%


\textbf{Correct proofs}
Correct proofs in PeanoBench are part of one of three groups --- (1) the \textit{staff-solutions}, the \textit{equation-based} proofs and the (2) \textit{justification-based} proofs.

The \emph{staff-solution} proofs are derived directly from NNG4 and annotated by two paper authors (before LeanTutor's architecture was designed), and the annotations are very descriptive. 
The two-other groups of proofs are generated according to ``personas''~\cite{cooper1999inmates}, a technique used in user-interface design as well as for synthetic data generation \citep{ge2024scaling}. 
The NL for the \emph{equation-based} persona consists primarily of algebraic manipulations and few words. The NL proof in the \emph{justification-based} persona explicitly contains the theorems and mathematical definitions used in the proof. Each proof was annotated by one of five (paper author) annotators and proofread by a different annotator. When possible, we varied the proof's Lean code (whether this be a major logical difference or rearranging commutative tactics) in addition to changing the NL comments according to the persona. Thus, we have a total of 225 correct proofs.


\subsubsection{Incorrect Proofs}
We additionally generate incorrect proofs by mimicking logical errors. 
Logical errors are a common mistake made by students, where they believe they have a valid proof, but it is in fact invalid
~\cite{weber2001student}. A common example for such a logical error is to assume a statement $\mathcal{P} \implies \mathcal{Q}$, but forgetting that there are steps required to justify this. We imitate this mistake and 
create incorrect proofs by randomly skipping a step from the last three lines of the proof 
(step-skipping algorithm pseudocode is in the Appendix).
Proofs that are only one line long are removed from the incorrect set.
Two limitations of this approach are that 1) this does not capture all possible error-types made by students, and 2) this programmatic deletion approach can sometimes lead to unrealistic errors.
In total, we end with $73 \times 2$ incorrect proofs in the \textit{equation-based} and \textit{justification-based} personas. We do not generate incorrect proofs using the \emph{staff-solution}.


\textit{Staff solutions} proofs are only offered as context to the model. System performance is evaluated on the correct and incorrect \textit{equation-based} and \textit{justification-based} proofs.

\section{System Design}
\begin{figure*}
\begin{center}
\includegraphics[width=1.0\linewidth]{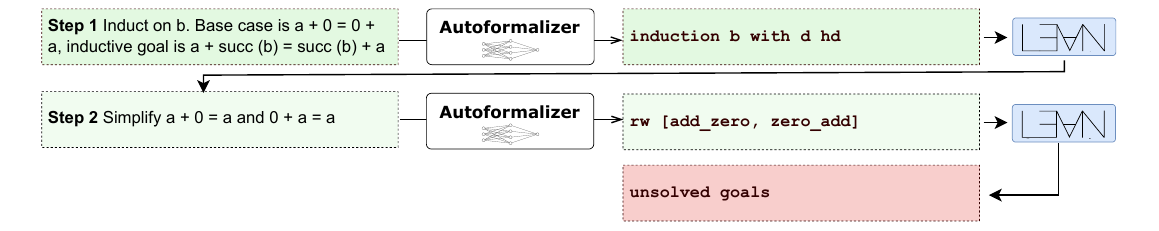}
\end{center}
\vspace{-0.5em}
\caption{Autoformalizer architecture. The natural langauge student step is provided to the autoformalizer, and the output is checked by the Lean compiler. The formalization of each step is appended to formalizations of previous steps to check for correctness. }
\label{fig:autoformalizer}
\vspace{-1em}
\end{figure*}

LeanTutor has three modules: an autoformalizer, a next-step generator, and an automatic feedback generator (Fig.~\ref{fig:systemdiagram}).

\subsection{Autoformalizer and Proof Checker}
\label{sec:autoformdesign}



%
 
%
%

We autoformalize student NL proofs one step at a time, and check for compliation at each step.  
This approach is similar in spirit to previous works breaking autoformalization into subtasks \citep{jiang2022draft}, but closest to the work of \citet{kulal2019spoc}.
Figure \ref{fig:autoformalizer} 
illustrates this process of translating a single student proof step into Lean, and repeating the process until the student is finished with their proof or the student makes an error in their proof. We prompt a general-purpose LLM to autoformalize the student proofs. 
Our prompts can be found in the Appendix.
(We do not pursue the post-training of or inference on open-source models at this time due to a severe limitation in both open-source models intended for whole proof autoformalization and the size of PeanoBench.) To support the autoformalization task, we add several key pieces of information in-context of our model:


\begin{itemize}
\item \textit{Staff Solution}: We provide one reference proof in both NL and FL. This reference proof may or may not align with the student proof being autoformalized.  We do not leverage this staff solution beyond providing it in the context, but aim to do so in future work. 
\item \textit{Theorem and Tactic Dictionary}: We create a dictionary of all theorems and tactics in the dataset,
where the keys are the formal Lean names of the theorems and tactics, and the values are natural language descriptions of each. 
\item \textit{5-shot examples}: We include five examples of translations of a natural language proof step and corresponding Lean formalization following~\citet{murphy2024autoformalizing}. 
\end{itemize}


We include informal proofs from PeanoBench and their corresponding faithful formalization in the Appendix. 


\subsubsection{Proof Checker}
The input to autoformalizer module will include both correct and incorrect proofs. As shown in Figure \ref{fig:autoformalizer}, each autoformalizing student proof step is appended to the Lean theorem statement and previously formalized steps. The proof is compiled, via LeanInteract \citep{leaninteract}. If the compiler output indicates only \texttt{unsolved goals}, we assume the student step is correct and proceed with autoformalizing remaining steps. For any other error message (\texttt{unknown tactic}, \texttt{error:unexpected identifier}, etc.), we assume the student step is incorrect and mark this proof step as erroneous. (Note: We end the autoformalization process once the first error is located.) Furthermore, a compiler error can indicate either an incorrect student proof step, or an autoformalization error. This is a limitation of the system that we intend to address in future work. 







\subsection{Next Step Generator}

The Next Step Generator (NSG) 
is launched when the student proof is not identified as complete and correct by the autoformalizer/proof checker. The NSG takes as input the formalized partial student proof (with the incorrect step removed).
It aims to output a  Lean tactic that can lead to a complete proof. Similar to \cite{thakur2023context}, the module performs an LLM-directed depth-first proof search. An LLM is instructed to generate 12 candidate tactics with a rank-ordering of their likelihood of being a correct next step. The prompt includes a list of all tactics/premises used in the NNG4 world of that theorem. 


The 12 generated tactic candidates are appended to the existing proof and run through the Lean compiler (via LeanInteract \citep{leaninteract}). Compiling tactics are then filtered through a \textit{progress check}, which follows~\citet{thakur2023context} and \citet{sanchez2020generating}. In the progress check, we (1) ensure we are not using any theorems on a list of forbidden theorems (we define this list to include the theorem we are currently trying to prove and theorems that are introduced after the theorem being proven in the order defined by NNG4) and (2) avoid cyclic tactics that would cause the proof-tree to revisit a goal state~\citep{thakur2023context}. We build a proof-search tree using all tactics that fulfill the compilation and progress check and do a depth-first search until a complete proof is found. We bound the tree depth to eight, which is sufficient for most of the proofs in our case. If a proof cannot be found, we report this to the following feedback generator module.


\subsection{Natural Language Feedback Generator}
The feedback-generation module combines information from previous modules to provide natural language feedback to the student. This module takes as input the student's autoformalized proof, the Lean compiler error message (if present), and the next Lean tactic generated from the NSG module. To aid in error identification, we include six common errors students have made in inductive proofs \citep{baker1996students} in our prompt (prompt in Appendix).

We use this information to automatically generate three types of feedback common in ITS. Similar to the automatic feedback generated by \citet{d2015can}, we (1) identify the student error 
and (2) generate a hint or question that guides the student to the next step. 
We also generate (3) an explicit next step the student could take, similar to a \textit{bottom-out hints}~\citep{suzuki2017exploring}.
This third part of our feedback is very similar to the auto-informalization task in automated theorem proving \citep{lisurvey}. 



\section{Experiments}
\begin{table*} [t]
  \caption{Autoformalization performance across correct and incorrect proofs. Autoformalization is done step-by-step in the Baseline and Baseline + Staff Solution experiments. In the experiments labeled with (whole proof), the whole NL proof is autoformalized into Lean code at once. Binomial error bars were computed using Jeffreys prior with a 95\% confidence interval.}
  \label{tab:autoformalizer-acc}
  \centering
  \begin{tabular}{lccc}
    \toprule
    Experiment & Correct Tactics & Correct Proofs & Incorrect Proofs \\
    \midrule
    Baseline & 32.9\% $\pm$ 3.1\% &  6.7\% $\pm$ 4.0\% & 14.4\% $\pm$ 5.7\% \\
    \textbf{Baseline + Staff Solution} &\textbf{ 56.8\% $\pm$ 3.2\%} & 18.0\% $\pm$ 6.1\% & \textbf{30.1\% $\pm$ 7.4\%} \\
    Baseline (whole proof) & 28.2\% $\pm$ 2.9\% & 10.7\% $\pm$ 4.9\% & 13.0\% $\pm$ 5.4\% \\
    Baseline + Staff Solution (whole proof) & 51.8\% $\pm$ 3.3\% & \textbf{26.7\% $\pm$ 7.0\%} & 21.9\% $\pm$ 6.7\% \\
    \bottomrule
  \end{tabular}
\end{table*}

We evaluate performance of the entire LeanTutor system on incorrect proofs. In this experiment, a baseline model and LeanTutor are both given incorrect proofs as input and generate NL feedback as output. Human evaluators then assess the generated feedback across four axes: Accuracy, Relevance, Readability, and Answer Leakage, on a 5-point scale. These experiments are detailed in section \ref{evals:system}. To understand the impact of key innovations in our autoformalizer, namely the presence of staff solutions and the step-by-step autoformalization approach, we perform experiments and ablations on just our Autoformalizer. To assess our model's performance at the faithful autoformalization task, we present a novel metric. These experiments are explained in section \ref{evals:autoformalizer}. All experiments cost less than \$4.00 to run on \texttt{gpt-4o-mini-2024-07-18}. We expect that the autoformalization performance can be boosted by using other more powerful LLMs. However, since the focus of our paper is presenting a model-agnostic framework for the LeanTutor model, we do not optimize over different LLMs.

\subsection{Metric for Faithful Autoformalization}
\label{sec:ourmetrics}
A few metrics have been developed to assess faithful autoformalization \citep{murphy2024autoformalizing, liurethinking, lin2025goedel, li2024autoformalize}. \citet{li2024autoformalize} verify Isabelle formalizations and rely on Sledgehammer, \citet{lin2025goedel} use an LLM-as-a-Judge, \citet{murphy2024autoformalizing} use an SMT solver to prove equivalence between two statements, and \citet{liurethinking} define a new equivalence relation. We prefer not to completely rely on the LLM-as-a-judge paradigm \citep{lin2025goedel} due to the potential for hallucinations. Both the measures proposed by \citet{murphy2024autoformalizing, liurethinking} are too coarse for our use case. 


We develop a metric that performs \textit{relaxed exact matching}. Our metric has two phases. Firstly, exact \textit{tactic-matching} is attempted in which the generated tactic string is matched with the ground truth tactic string.
If string matching fails, we move to the second phase: \textit{state-matching}. In \textit{state-matching} we compare the two tactics by checking if the proof states (the proof state rendered once the predicted and ground truth tactics have been appended to the existing predicted and ground truth proofs respectively) are syntactically identical up to variable naming. We call our metric \textit{relaxed}, because we accommodate differing variable names between the input and ground truth proofs. To do this, proof states are segmented by goal and/or casework and we locate all variables through a custom Python implementation of Lean Identifiers \citep{lean_syntax}. Variables in all goal state segments are standardized and string matching can ensue. If this check fails as well, we deem the predicted tactic as not a faithful autoformalization of the input NL proof stem. More details on metric implementation can be found in the Appendix. 

\subsection{Autoformalizer Evaluation}
\label{evals:autoformalizer}

Our \textit{baseline model} adapts the autoformalization prompt proposed by \citet{murphy2024autoformalizing} to our dataset. Their autoformalization prompt was designed for a small dataset use case in which all tactics/premises can be provided in-context; this is appropriate for PeanoBench.
Our baseline prompt
contains the theorem statement in both NL and FL, the tactic and theorem dictionaries, five examples of the formalization task, and the student input that needs to be formalized. 

All model outputs are evaluated at \textit{pass@1}, in contrast with some prior work in autoformalization in which performance has been evaluated at pass@8 \citep{liurethinking}. 
For correct proof formalizations, accuracies at both the tactic and proof levels were measured. Tactic-level accuracies were determined using the metric described above. Proof-level accuracy was measured by verifying that all NL statements in a given proof were autoformalized into the correct tactics, in the correct order. Thus, proof-level accuracy measures a much higher performance bar than does tactic-level accuracy. For incorrect proof formalizations, we report only proof-level accuracy. Thus, for incorrect proofs a formalization is considered successful if (1) all correct proof steps until the first incorrect step were formalized correctly and (2) formalization of the incorrect proof step leads to a Lean compiler error.


\subsubsection{Findings} We report results in Table \ref{tab:autoformalizer-acc}. Tactic-level results are out of 900 total tactics, correct proofs results are out of 150 total proofs, and incorrect proof results are out of 146 proofs. The Baseline + Staff Solution model displays superior performance in all categories compared to the baseline. 

 We find that the model relies heavily on the staff solution in its context in generating the formalization. In 89\% of correct tactic formalizations, the tactic predicted was in the staff solution (meaning that the tactic was present somewhere in the staff solution provided in context, but possibly in a different position than the predicted tactic). 
When the autoformalization was wrong, the model copied a tactic in the staff-solution, instead of predicting the real tactic corresponding to the NL input, in 51\% of cases. When the expected formalization was not in the staff solution, LeanTutor was able to correctly formalize the NL proof step in 32\% of cases. 
These results demonstrate a limitation of the current approach's ability to autoformalize NL proofs whose formalization does not correspond to the formalized staff solution provided in-context. 


\subsubsection{Ablations} We compare our autoformalizer model to one ablation: generating whole proofs all at once instead of step-by-step generations (experiments labeled with (whole proof) in Table \ref{tab:autoformalizer-acc}). 
With this approach, the autoformalized whole proof does not necessarily contain the same number of tactics as our ground truth whole proof. We truncate proof lengths to \texttt{min(len(generated proof), len(ground truth proof))} (the length of a proof referring to the number of tactics in the proof) and align both proofs to each other tactic-by-tactic. We compute tactic-level and proof-level accuracy in the same manner described above\footnote{Our metric is imperfect for evaluating generated whole proofs. Thus, we also evaluate how many generated whole proofs (in the correct proof experiments) also completed successfully, with the Lean compiler displaying \texttt{no goals}. The Baseline (whole proof) model produced 28 compiling proofs and the Baseline + Staff Solution model (whole proof) produced 50 compiling proofs. Note, that a complete Lean proof doesn't serve as an appropriate measure for faithful autoformalization.}. Considering the models with staff solutions, the step-by-step autoformalization approach has comparable performance to the whole proof autoformalization on correct proofs. However, the step-by-step approach outperforms the whole proof approach on incorrect proofs, by  8\%. As many incoming proofs to a tutoring system will be incorrect, better performance on incorrect proofs is advantageous. 

Prompts for step-by-step and whole proof generation can be found in the Appendix. 
We performed additional experiments, evaluating the impact of adding the student's natural language proof and Lean goal state information in-context of the autoformalizer. However, these experiments performed lower than the Baseline + Staff Solution experiment. These results can be found in the Appendix. 

\begin{table*}[ht]
    \centering
    \begin{tabular}{lllll}
     \toprule
        \multicolumn{1}{c}{\bf Feedback Type} & \multicolumn{1}{c}{\bf Accuracy} & \multicolumn{1}{c}{\bf Relevance} & \multicolumn{1}{c}{\bf Readability} & \multicolumn{1}{c}{\bf Answer Leakage} \\
        \midrule
        Baseline Error Identification & \textbf{3.4} & 3.5 & \textbf{4.5} & \textbf{4.6} \\
        LeanTutor Error Identification & \textbf{3.4} & \textbf{3.8} & 4.3 & 4.5 \\
        \midrule
        Baseline Hint/Question & 3.1 & 2.9 & \textbf{4.7} & 4.3 \\
        LeanTutor Hint/Question & \textbf{3.7} & \textbf{3.7} & \textbf{4.7} & \textbf{4.4} \\
        \midrule
        Baseline Next Step & 2.8 & 2.9 & \textbf{4.6} & \textbf{2.2} \\
        LeanTutor Next Step & \textbf{3.7} & \textbf{3.9} & 4.4 & 1.1 \\
        
        \bottomrule
    \end{tabular}
    \caption{Average (across all proofs) scores of generated feedback from baseline and LeanTutor experiments on 21 incorrect proofs. The generated feedback was scored on a scale of 1-5 in which a score closer to 5 indicates desired performance.}
    \label{tab:system-eval}
\end{table*}

\subsection{Metric for LeanTutor Feedback}
In the system-level evaluation of LeanTutor, a student NL proof is input and NL feedback is generated as output. We evaluate the generated outputs on four axes: \textit{Accuracy}, \textit{Relevance}, \textit{Readability}, and \textit{Answer Leakage}, motivated by the metrics used in~\citet{mitra2024retllm, mozafari2025hinteval}. 
We evaluate each of our three categories of feedback (error identification, hint/question generation and explicit next step) along each axis using a 5-point scale.

We define what it means to receive the highest rating of 5 for each axis. A score of 1 indicates complete disagreement with the following definitions. \textbf{Accuracy}: The generated error/hint/next-step is correctly and accurately identified \citep{mitra2024retllm}. \textbf{Relevance}: The generated error/hint/next step is relevant to the error/proof following ~\citet{mitra2024retllm, mozafari2025hinteval}. \textbf{Readability}: The generated feedback is coherent~\citep{mitra2024retllm}. \textbf{Answer Leakage}: The generated feedback does not disclose the answer in any way~\citep{mozafari2025hinteval}. 




\subsection{LeanTutor Evaluation}
\label{evals:system}

We evaluate our full system on incorrect proofs and ``cold-start'' proofs, a proof in which the student does not know how to start the proof. 
Results for the ``cold-start'' proofs can be found in the Appendix. 
Across our experiments, we use \texttt{gpt-4o-mini-2024-07-18} (temperature $= 0.0$). Two external evaluators (graduate students in the Computer Science department with prior teaching experience in math courses) scored all generated feedback from the baseline and LeanTutor models. Both evaluators were provided the same evaluation protocol document and had a brief discussion with a paper author, to explain the scoring task. The evaluations were model-blind---the evaluators did not which model generated which feedback. Each evaluator scored half the proofs, so each proof's feedback was evaluated by a single evaluator. 

We evaluate our end-to-end system on a subset of incorrect proofs from PeanoBench. We only consider incorrect proofs that were ``successfully autoformalized'' by the LeanTutor autoformalizer. Of the 44 proofs (results in Table \ref{tab:autoformalizer-acc}), we randomly selected one to three proofs per world, totaling 21 proofs\footnote{We excluded proofs where the skipped step did not lead to a Lean compiler error.}. 
These proofs are passed through our Next Step Generator and Feedback Generator modules. We compare to a simple baseline, providing the LLM with the erroneous student proof and prompting the model to generate the three feedback types. 
All prompts can be found in the Appendix.
An example of LeanTutor's generated hints (and their evaluations) for one incorrect proof are also in the Appendix.

Our system-level evaluation (Table \ref{tab:system-eval}) indicates LeanTutor outperforms the baseline model on the \textit{Accuracy} and \textit{Relevance} metrics. Performance on the \textit{Readability} and \textit{Answer Leakage} metrics are comparable for both models. (Note: We expect high answer leakage in the scores for ``next step'' feedback; a score of 1 is expected).

We also conduct the same evaluation, with \texttt{gpt-4} as the judge, following the ``LLM-as-a-judge'' paradigm. The ``judge'' was calibrated to three proofs' feedback scores given by the paper authors. 
The results (in the Appendix) show that the baseline model edges out LeanTutor in almost every scoring axis on every feedback type. 
However, these results do not align with our human evaluators' scores, leading us to ultimately not consider them.












\section{Limitations}

LeanTutor presents a proof-of-concept design for a formally-verified mathematical proof tutoring system, and leaves many open questions for future research. Some limitations come from assumptions in our system design, that may make it difficult to generalize our system.
(1) Assuming a one-to-one correspondence between NL proof steps and FL tactics, which generally will not scale to more complicated proofs. (2) We assume the presence of an already formalized staff solution, which could be a significant burden on an instructor in the absence of a good autoformalizer. (3) Our metric for faithful autoformalization applies only when ground truth formalizations exist. We aim to explore approaches that only use an informal staff solution. 

A second set of limitations comes from our 
dataset construction. (1) While students commonly miss steps in writing proofs, there are other types of errors that are not captured in the incorrect proofs dataset. (2) All of the natural language in PeanoBench has been written by paper authors, as opposed to non-author students (the varied personas are an earnest effort to incorporate realistic natural language variations).

Finally, a critical limitation in our system design is that if autoformalization fails, we may provide incorrect feedback or not be able to respond to the student.  
As a result, in our evaluation, we did not evaluate end-to-end system performance on proofs with incorrect autoformalization. 
Relatedly, we assume a student proof is incorrect if the Lean compiler errors. However, errors may also result from incorrect autoformalization, which could lead to false positives (though spot checking revealed this was not a big issue).

\section{Conclusions and Future Work}

Our future goal is to deploy LeanTutor in large undergraduate mathematics classes such as discrete math and linear algebra. Our next-steps include: 
\begin{itemize}
    \item more effective use of formal/informal reference proofs in the Autoformalizer and NSG.
    \item implement backtracking in the next-step generation module to discard unviable student steps and generate a feasible next step, starting from the most recent point in the student's proof that can lead to a complete proof.
\end{itemize}
Additionally, we have the open challenges on faithful autoformalization due to granularity and reasoning paths as discussed in the introduction. 
%
%
Finally, the tutoring application really forces a focus on the \emph{human} factors that must go into joint human-AI systems. While the current paper does not focus on these factors, it opens the door to working on them. Our hope is that LeanTutor's approach of combining state-of-the-art LLMs with the Lean theorem prover inspires future systems that combine LLMs with external verifiers in educational applications.

\section*{Acknowledgements}
The authors would like to acknowledge support from: NSF
CAREER grant ECCS-2240031, the 2023 CITRIS Institute
Seed Grant, and the UC Berkeley Instructional Technology
and Innovation MicroGrant Program.

\appendix

\lstset{%
	basicstyle={\footnotesize\ttfamily},
	numbers=left,numberstyle=\footnotesize,xleftmargin=2em,
	aboveskip=0pt,belowskip=0pt,%
	showstringspaces=false,tabsize=2,breaklines=true}
\floatstyle{ruled}
\newfloat{listing}{tb}{lst}{}
\floatname{listing}{Listing}
%
\pdfinfo{
/TemplateVersion (2026.1)
}

\setcounter{secnumdepth}{2} 


\lstdefinestyle{leanstyle}{
  language=lean,
  basicstyle=\ttfamily\small,
  keywordstyle=\bfseries\color{blue},
  commentstyle=\itshape\color{gray},
  breaklines=true,
  columns=fullflexible,
  keepspaces=true,
  showstringspaces=false,
  frame=single,
  framerule=0.8pt,
  backgroundcolor=\color{gray!5},
  captionpos=b
}
\lstset{
  basicstyle=\ttfamily\footnotesize,
  numbers=left,
  numberstyle=\tiny,
  stepnumber=1,
  numbersep=8pt,
  commentstyle=\color{gray!70!black}\itshape,
  keywordstyle=\bfseries,
  columns=fullflexible,
  keepspaces=true,
  showstringspaces=false,
  frame=single,
  breaklines=true,
  breakindent=0pt,
  tabsize=2,
  }

\appendix
\section{Extended Review of Math Proof Tutors}
\label{appendix:lit-review-tutors}

We identify three main categories of autonomous proof tutoring systems: (1) intelligent tutoring systems, (2) LLM-based tutoring systems, and (3) theorem prover based systems. Each of these systems has unique advantages, which LeanTutor attempts to build upon. 

\subsection{Intelligent Tutoring Systems.} \citet{corbett1997intelligent} characterize a system as an \textit{intelligent tutoring system} (ITS) if it fulfills eight design principles, which include: scaffolding student learning, modeling students' learning trajectories over time, and providing immediate feedback. Researchers have made attempts to develop \citep{autexier2012towards, briggle2008towards} or developed ITS for math proofs \citep{barnes2008toward, lodder2021generation, bundy2000intelligent}. ITS maintain a high quality of education through expert authoring of solutions or feedback, but this also makes them difficult to develop and scale \cite{dermeval2018authoring}. To reduce this burden, LeanTutor dynamically generates proof trees based on student solutions, similar to \citet{lodder2021generation} approach, but in contrast, also generates feedback on-demand via a generative language model. 
\\
\subsection{LLM-Based Tutors.} Given the extremely recent advance of high performance LLMs, there are not yet many LLM-based math tutors for proofs specifically. \citet{zhao2024autograding} propose an LLM-based autograder for inductive proofs, which provides students with real-time feedback on the correctness of their proofs. \citet{park2024using} evaluated ChatGPT's abilities to aid students in refining and improving their proofs. Broadly speaking, many LLM-based math tutors have been developed and studied \citep{tonga2024automatic, miller2024llm, autexier2012towards,wang2024tutor, park2024using}. These math tutors have shown to maintain conversations without inappropriate content \citep{levonian2025designing} and even lead to learning gains for students studying algebra \citep{pardos2023learning}. However, LLMs still cannot suffice as effective tutors due to (1) hallucinations,\citep{maurya2024unifying, balunovic2025mathconstruct} (2) models revealing the whole answer \citep{sonkar2024pedagogical}, (3) models do not necessarily provide the correct reasoning behind an answer \citep{gupta2025beyond}, and (4) models struggle to identify mistakes \citep{tyen2024llms, miller2024llm}. LeanTutor capitalizes on the conversational ability of LLMs, but ``outsources'' reasoning tasks to theorem provers.

\subsection{Proof Assistant-based Tutors.} Theorem provers, such as Lean \citep{moura2021lean}, Coq \citep{huet1997coq}, and Isabelle \citep{paulson1994isabelle}, have all been used by some math educators as tools to teach students proofs \citep{avigad2019learning, villadsen2021using, boldo2024teaching, kerjean2024maths}. Additionally, proof tutors or educationally-geared tools have been developed on top of theorem provers: ProofTutor using APRoS \citep{sieg2007apros}, ProofWeb \citep{hendriks2010teaching} based on Coq, JAPE \citep{sufrin1997jnj}, Waterproof \citep{wemmenhove2022waterproof} built on Coq, HazelProver built on Agda \citep{omar2019live, keenanlearner}, Verbose Lean based on Lean \citep{massot2024teaching}, and MathsTiles build on Isabelle/HOL \citep{billingsley2007student}. These tools have led to unique benefits in students' learning of proofs \citep{thoma2022learning}, but students struggle to learn the complex syntax required to interact with most \citep{avigad2019learning, buzzard_teaching, villadsen2021using, karsten2023proofbuddy}. LeanTutor combats this issue by allowing the student to interface only in natural language and hiding the Lean formalizations of student proofs altogether.

\section{Proofs from PeanoBench}
\label{appendix:proofexamples}
The PeanoBench dataset contains three main subsets of proofs: \textit{staff solution} proofs, \textit{correct}  proofs, and \textit{incorrect} proofs. \textit{Correct} proofs are derived from the staff solution proofs, with two main differences: (1) Lean syntax in the proof is changed when possible and (2) the NL in-line comments are in differing ``personas'' (the equation-based and justification-based personas). Figure \ref{fig:correctproofsaddcomm} demonstrates the \textit{staff solution} proof of the theorem \texttt{add\_comm} (proving the commutativity of addition) as well as the equation-based and justification-based commented versions of the original proof (with small changes in Lean code). Figure \ref{fig:incorrectproofaddcomm} is an example of an incorrect proof of \texttt{add\_comm}, created by skipping a step in the justification-based persona proof. 

\begin{figure*}

\centering
\begin{minipage}{\textwidth}
\begin{lstlisting}[style=leanstyle, escapeinside={(*}{*)}, lineskip=-1.5pt]
theorem add_comm_staff_solution (a b : (*$\mathbb{N}$*)) : a + b = b + a := by
  -- Induct on b, with d = 0 as the base case and the inductive hypothesis a + d = d + a. There are now two proof goals, prove base case: a + 0 = 0 + a and the inductive step: a + succ d = succ d + a
  induction b with d hd
  -- First prove base case. Simplify LHS a + 0 to a.
  rw [add_zero]
  -- Simplify RHS 0 + a to a
  rw [zero_add]
  -- Prove LHS and RHS are equal, a = a, completing the base case.
  rfl
  -- Now prove the inductive step. Rewrite LHS a + succ (d) to succ (a + d)
  rw [add_succ]
  -- Rewrite RHS succ (d) + a to succ (d + a)
  rw [succ_add]
  -- Rewrite LHS succ (a + d) to succ (d + a) using the inductive hypothesis
  rw [hd]
  -- Prove succ LHS and RHS are equal, (d + a) = succ (d + a), completing the proof
  rfl
\end{lstlisting}
\end{minipage}

\begin{minipage}{\textwidth}
\begin{lstlisting}[style=leanstyle, escapeinside={(*}{*)}, lineskip=-1.5pt]
theorem add_comm_equation_based (a b : (*$\mathbb{N}$*)) : a + b = b + a := by
  -- Start by inducting on b
  induction b with d hd
  -- 0 + a -> a on RHS giving us a + 0 = a
  rw [zero_add]
  -- a + 0 -> a into the LHS to get a = a
  rw [add_zero]
  -- a=a, we are done with the base case
  rfl
  -- a + succ d -> succ (a + d) on LHS giving us succ (a + d) = succ d + a
  rw [add_succ]
  -- succ d + a -> succ (d + a) on RHS giving us succ (a + d) = succ (d + a)
  rw [succ_add]
  -- using the induction hypothesis, succ (a + d) -> succ (d + a) on the LHS giving us succ (d + a) = succ (d + a)
  rw [hd]
  -- succ (d + a) = succ (d + a), we are done.
  rfl
\end{lstlisting}
\end{minipage}

\begin{minipage}{\textwidth}
\begin{lstlisting}[style=leanstyle, escapeinside={(*}{*)}, lineskip=-1.5pt]
theorem add_comm_justification_based (a b : (*$\mathbb{N}$*)) : a + b = b + a := by
  -- Start by inducting on b
  induction b with d hd
  -- We start with the base case. using properties of addition by 0 we can rewrite a + 0 to a on the LHS
  rw [add_zero]
  -- using properties of addition by 0 we can rewrite 0 + a to a on the RHS
  rw [zero_add]
  -- since both sides are equal, we are done with the base case
  rfl
  -- Now to the (n+1) step. using properties of successors, succ (n) + a -> succ (n + a) and substitute this into the RHS
  rw [succ_add]
  -- using properties of succession, we substitute a + succ(n) -> succ(a+n) on the RHS
  rw [add_succ]
  -- Use the induction hypothesis on the LHS to substitute succ (a + n) -> succ (n + a)
  rw [hd]
  -- since both sides are equal, we are done with the proof
  rfl
\end{lstlisting}

\end{minipage}
\caption{Examples of annotated Peano Arithmetic proofs from PeanoBench for the theorem proving commutativity of addition, that is, for all \( a, b \in \mathbb{N} \), \(a + b = b + a\). The first proof,\texttt{add\_comm\_staff\_solution} follows the exact Lean code from NNG4. The second and third proofs, \texttt{add\_comm\_equation\_based} and \texttt{add\_comm\_justification\_based}, are written in two different personas.}
\label{fig:correctproofsaddcomm}

\end{figure*}

\begin{figure*}
\begin{lstlisting}[style=leanstyle, escapeinside={(*}{*)}, lineskip=-1.5pt]
theorem add_comm_incorrect (a b : (*$\mathbb{N}$*)) : a + b = b + a := by
  -- Start by inducting on b
  induction b with d hd
  -- We start with the base case using properties of addition by 0 we can rewrite a + 0 to a on the LHS
  rw [add_zero]
  -- using properties of addition by 0 we can rewrite 0 + a to a on the RHS
  rw [zero_add]
  -- since both sides are equal, we are done with the base case
  rfl
  -- Now to the (n+1) step. using properties of successors, succ (n) + a -> succ (n + a) and substitute this into the RHS
  rw [succ_add]
  -- using properties of succession, we substitute a + succ(n) -> succ(a+n) on the RHS
  rw [add_succ]
  -- since both sides are equal, we are done with the proof
  rfl
\end{lstlisting}

\caption{Example of an incorrect proof for the theorem proving commutativity of addition, that is, for all \( a, b \in \mathbb{N} \), \(a + b = b + a\). This proof, originally the justification-based persona, has the \texttt{rw [hd]} step, which applies the inductive hypothesis, skipped.} 
\label{fig:incorrectproofaddcomm}
\end{figure*}

\newpage
\section{Proof Breakdown by Worlds}
\label{appendix:worldbreakdown}
NNG4 categorizes proofs based on distinct worlds. The table below presents the distribution of proofs across these worlds, illustrating the relative frequency of each category.

\begin{table}
    \centering
    \begin{tabular}{lll}
    \toprule
        \multicolumn{1}{c}{\bf World} & \multicolumn{1}{c}{\bf \# Tactics} & \multicolumn{1}{c}{\bf \# Proofs} \\
        \midrule
        Implication & 38 & 13 \\ 
        Multiplication & 57 & 9 \\ 
        Advanced Multiplication & 66 & 10 \\ 
        Algorithm & 20 & 5 \\ 
        Less or Equal & 86 & 11 \\ 
        Power & 70 & 9 \\ 
        Tutorial & 24 & 7 \\ 
        Advanced Addition & 32 & 6 \\ 
        Addition & 41 & 5 \\ 
        \midrule
        \textbf{Total} & \textbf{434} & \textbf{75} \\ 
        \bottomrule
    \end{tabular}
    \\
    \caption{Distribution of selected proofs from NNG4 by world.}\label{solution-proof-counts}
\end{table}

\section{PeanoBench: Incorrect Proof Generation Algorithm}
\label{sec:stepskipping}

\begin{algorithm}[H]
\caption{\textsc{Step-skipping algorithm for generating incorrect proofs.}}
\label{alg:stepskipping}
\renewcommand{\thealgorithm}{Procedure A}
\begin{algorithmic}
\FORALL{$P \in \textsc{CorrectDeviatingProofs}$}
    \STATE $n \gets \text{length}(P)$
    \IF{$n = 2 \textbf{ or } n = 3$}
        \STATE delete step 2
    \ELSIF{$n = 4$}
        \STATE randomly delete step $n - 1$ or $n - 2$
    \ELSIF{$n > 4$}
        \STATE randomly delete one of step $n - 1$, $n - 2$, or $n - 3$
    \ENDIF
\ENDFOR
\end{algorithmic}
\end{algorithm}

\section{Next Step Generator}
\begin{figure*}
\begin{center}
\includegraphics[width=1.0\linewidth]{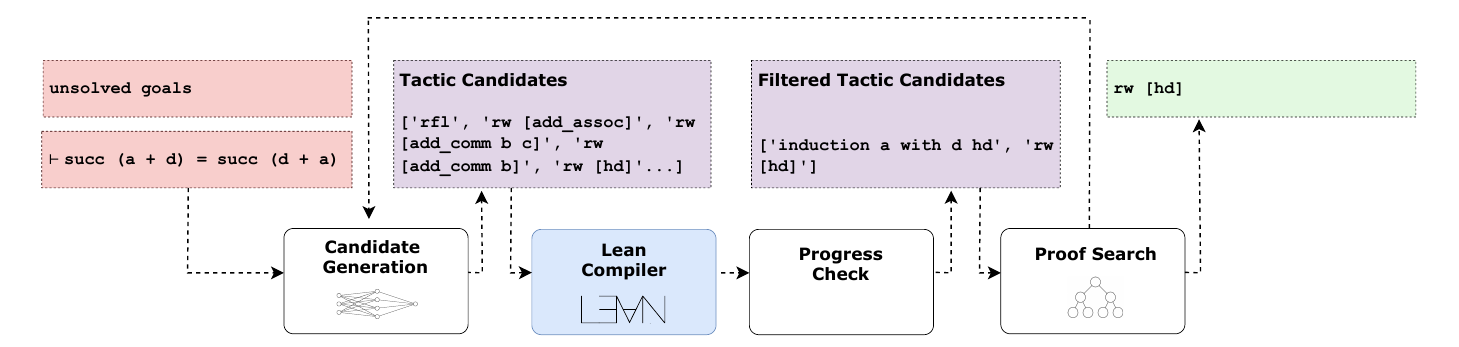}
\end{center}

\caption{Architecture of the Next Step Generation module. An LLM generates tactic candidates which are appended to the pre-existing proof. Tactics that compile correctly are then passed through a ``progress check'' filter which ensures goal states are not being re-visited. This process of generating and checking tactics is repeated until the proof is completed.}
\label{fig:nsg}

\end{figure*}

The architecture for the next step generator is in Figure \ref{fig:nsg}
\label{appendix:nsg}

\section{Autoformalizer Extended Results}
\label{appendix:autoformexps}
We additionally experiment with adding the following information into the autoformalizer prompt. All formalizations were generated step-by-step (Section \ref{sec:autoformdesign}). 

Experiments include: 
\begin{itemize}
    \item \textit{Staff Solution}: The staff solution proof, a complete and correct proof for the theorem in both NL and FL. The autoformalizer accuracy with the staff solution is also presented in the main paper.
    \item \textit{Previous NL}: The student's previous proof steps (in natural language) up until that point, 
    \item \textit{Previous Goal State}: The Lean goal state of the proof formed by appending autoformalizations of the student's NL proof to the Lean theorem statement. (Note that this goal state may become ``corrupted'' if any previous formalizations were incorrect. If a goal state displays an error message, we did not include the goal state in the prompt and the prompt was then identical to the baseline.)
\end{itemize}

The results of these experiments (in addition to experiments discussed in the main paper) are summarized in Table \ref{tab:autoformexps}.

\begin{table*}
  \label{tab:proof-accuracy}
  \centering
  \begin{tabular}{lccc}
    \toprule
    Condition & Correct Tactics & Correct Proofs & Incorrect Proofs \\
    \midrule
    Baseline & 296 / 900 = 32.89\% & 10 / 150 = 6.67\% & 21 / 146 = 14.38\% \\
    + Staff Solution & 511 / 900 = 56.78\% & 27 / 150 = 18.00\% & 44 / 146 = 30.14\% \\
    + Previous Goal State & 312 / 900 = 34.67\% & 15 / 150 = 10.00\% & 29 / 146 = 19.86\% \\
    + Previous NL & 331 / 900 = 36.78\% & 10 / 150 = 6.67\% & 20 / 146 = 13.70\% \\
    + Previous NL + Staff Solution & 522 / 900 = 58.00\% & 28 / 150 = 18.67\% & 42 / 146 = 28.77\% \\
    Whole Proof (Baseline) & 254 / 900 = 28.22\% & 16 / 150 = 10.67\% & 19 / 146 = 13.01\% \\
    + Whole Proof (Staff Solution) & 466 / 900 = 51.78\% & 40 / 150 = 26.67\% & 32 / 146 = 21.92\% \\
    \bottomrule
  \end{tabular}
  \caption{Extended autoformalizer experiment results.}
  \label{tab:autoformexps}
\end{table*}

\section{Metric}
\label{appendix:metric}
Since we are interested in faithful autoformalization, we measure the accuracy of our autoformalizer on a tactic-by-tactic basis. For this, we check that check either the tactic itself or the proof state after every tactic matches the corresponding ground truth tactic/proof state. First, the tactics themselves are compared using exact string matching, with the minor exception that \texttt{rw [...} and \texttt{rw[...} (the only difference between the strings is the space before the brackets) are considered equivalent.
This covers a lot of cases, but sometimes two tactics behave identically, but are not literally the same string (for example, \texttt{rw [add\_comm]} and \texttt{rw [add\_comm a b]} might do the same thing in a proof, but string matching would fail). Additionally, two tactics might use different variable names (for example, \texttt{induction n with d hd} and \texttt{induction n with k hk} are equally valid). So, we cannot just use exact string matching.

If string matching does not identify the tactics as identical, then the tactics are verified in Lean (appended to any previous tactics for the predicted and ground truth proofs respectively) and we check if the resulting proof states are syntactically identical up to variable naming. If either the string matching or proof state matching check succeeds, the generated tactic is considered correct. 
By ``up to variable naming'', we mean that two goals are considered equivalent if they are structurally the same, but may use different variable names. For example, the following proof states are identical up to variable naming, but neither of them are exact string matches.

\begin{lstlisting}[escapeinside={(*}{*)}]
n : (*$\mathbb{N}$*)
h : 1 (*$\le$*) n
(*$\vdash$*) n + 0 = n
\end{lstlisting}

\begin{lstlisting}[escapeinside={(*}{*)}]
m : (*$\mathbb{N}$*)
hm : 1 (*$\le$*) m
(*$\vdash$*) m + 0 = m
\end{lstlisting}


\subsection{Proof State Comparison}
The algorithm to compare proof states up to variable renaming works as follows. First, the proof states are split into cases and each case is compared individually. All cases must be equivalent for the proof states to be considered equivalent. Then, within each case, free variables (which are not bound by a binder and can be seen for the first time above the $\vdash$) \footnote{Lean supports three types of variables: bound variables, which first appear under a \emph{binder} such as $\forall$ or \texttt{fun}; free variables, which are not bound by a binder and can be seen for the first time above the $\vdash$; and meta-variables, which represent holes in an expression that must be filled in before the proof is complete. Only free variables are supported for variable renaming; bound variables and meta-variables are not handled because they rarely ever appear within proof states in NNG4 and handling them would amount to a drastic increase in complexity.} are identified by checking what appears before the first colon on each line. In the proof states below, \texttt{n} and \texttt{hn} in the first proof and \texttt{m} and \texttt{hm} in the second proof are all free variables. After identifying free variables, the proof states are normalized by renaming each appearance of a variable according to its position in the variable list (see Algorithm \ref{alg:normproofstate}).
\footnote{
Our code to determine what constitutes a valid Lean identifier does not handle double guillemets 
because they are not used in NNG4.}

The proof state normalization algorithm is written in Algorithm \ref{alg:normproofstate}. To normalize a proof case (one case in a proof state), we make a list of all variables (including proofs) in the local context, which includes everything listed before a colon in a line above the $\vdash$. Next, we locate all identifiers in the goal states we are comparing via a Python implementation of Lean identifiers \citep{lean_syntax}. An identifier in Lean is a string that acts as a variable name or refers to a constant such as a theorem or a type.
For example, \texttt{x} and \texttt{MyNat.add\_comm} are both identifiers. Identifiers that match a variable name are replaced with \texttt{var\emph{i}}, where \emph{i} is the index of the variable in the variable list created earlier. 
To locate identifiers, we use a greedy algorithm which loops through all characters in the proof state.

\newpage
So, for example, the following proof states, 
\begin{lstlisting}[escapeinside={(*}{*)}]
n : (*$\mathbb{N}$*)
h : 1 (*$\le$*) n
(*$\vdash$*) n + 0 = n
\end{lstlisting}

\begin{lstlisting}[escapeinside={(*}{*)}]
m : (*$\mathbb{N}$*)
hm : 1 (*$\le$*) m
(*$\vdash$*) m + 0 = m
\end{lstlisting}
would both be converted to
\begin{lstlisting}[escapeinside={(*}{*)}]
var0 : (*$\mathbb{N}$*)
var1 : 1 (*$\le$*) n
(*$\vdash$*) var0 + 0 = var0
\end{lstlisting}

The algorithm \textit{Normalize} is below. Note that \textit{GetVariables} is a function that collects all the variables from proof state as described earlier.

\begin{algorithm}[H]
\caption{Normalize Proof State}
\begin{algorithmic}[1]
\STATE \textbf{function} Normalize(proof\_state)
    \STATE variable\_list $\leftarrow$ GetVariables(proof\_state)
    \STATE result $\gets$ ""
    \STATE $i \gets 0$
    \WHILE{$i <$ len(proof\_state)}
        \STATE ident $\leftarrow$ LongestIdentifierStartingAt(proof\_state, $i$)
        \IF{ident $\ne$ Null}
            \IF{ident $\in$ variable\_list}
                \STATE result $\leftarrow$ result $+$ "var" $+$ IndexOf(ident, variable\_list)
            \ELSE
                \STATE result $\gets$ result $+$ ident
            \ENDIF
            \STATE $i \gets i +$ len(ident)
        \ELSE
            \STATE result $\leftarrow$ result $+$ GetChar(proof\_state, $i$)
            \STATE $i \gets i + 1$
        \ENDIF
    \ENDWHILE
    \STATE \textbf{return} result

    
\end{algorithmic}
\label{alg:normproofstate}
\end{algorithm}

The \textit{Normalize} algorithm relies on the \textit{LongestIdentifierStartingAt} algorithm as described below.
\begin{algorithm}[H]
\caption{Longest Identifier}
\begin{algorithmic}[1]
\label{alg:longestid}
\STATE \textbf{function} LongestIdentifierStartingAt(str, $i$)
\STATE len $\leftarrow 0$
\IF{IsValidLeanIdentifier(Substring(str, $i$, $i + 2$))}
    \STATE len $\leftarrow 2$
\ENDIF
\WHILE{IsValidLeanIdentifier(Substring(str, $i$, $i + \text{len} + 1$)) \AND $i + \text{len} <$ Len(str)}
    \STATE len $\leftarrow$ len $+ 1$
\ENDWHILE
\IF{len $>$ 0}
    \STATE \textbf{return} Substring(str, $i$, $i + \text{len}$)
\ELSE
    \STATE \textbf{return} Null
\ENDIF

\end{algorithmic}
\label{alg:longest}
\end{algorithm}

\section{Examples of Generated Hints}
\label{appendix:hintexample}
We demonstrate an example of the three generated hint types (error identification, question, bottom-out hint) for the following theorem. The incorrect proof skips the \texttt{tauto} tactic and results in the base case never being proven. Unfortunately, LeanTutor's feedback fails to get at the heart of the student's issue and seems more inclined towards the inductive case.

\begin{figure}
\vspace{-1.5em}
\centering
\begin{lstlisting}[style=leanstyle, escapeinside={(*}{*)}, lineskip=-1.5pt]
-- Theorem Statement: Prove that if a is not equal to 0, then a is the successor of some natural number.
theorem eq_succ_of_ne_zero_correct (a : (*$\mathbb{N}$*)) (ha : a  0) : (*$\exists$*) n, a = succ n := by
  -- Induct on a
  induction a with d _
  -- For the base case, a = 0, we have a contradiction in hypotheses because we know a (*$\neq$*) 0.
  tauto
  -- For the inductive step, we set n to be d.
  use d
  -- succ d = succ d
  rfl
\end{lstlisting}

\begin{lstlisting}[style=leanstyle, escapeinside={(*}{*)}, lineskip=-1.5pt]
-- Theorem Statement: Prove that if a is not equal to 0, then a is the successor of some natural number.
  theorem eq_succ_of_ne_zero_incorrect (a : (*$\mathbb{N}$*)) (ha : a (*$\neq$*) 0) : (*$\exists$*) n, a = succ n := by
  -- We use induction on a.
  induction a with d _
  -- For the inductive step, we set n to be d which gives us the goal succ d = succ d
  use d -- incorrect
  -- Since the LHS and RHS are equal, we can use reflexivity to prove the goal.
  rfl -- incorrect
\end{lstlisting}

\end{figure}

\textbf{1. Error Identification}: You are incorrectly inducting on the variable 'a' instead of its successor form, which leads to confusion in your proof.

[Accuracy: 1, Relevance: 1, Readability: 4, Answer Leakage: 5]

\textbf{2. Question}: Do you think you should consider the structure of 'a' in terms of its base case and successor?

[Accuracy: 3, Relevance: 2, Readability: 3, Answer Leakage: 4]

\textbf{3. Bottom-out Hint}: The next step is to use the fact that if 'a' is a successor, you should analyze 'a' as 'succ d' for some natural number 'd'.

[Accuracy: 4, Relevance: 3, Readability: 5, Answer Leakage: 1]

\section{Cold-start Proof Results}
\label{appendix:coldstart}

 A ``cold-start'' proof is a proof in which the student does not know how to start the proof. For this experiment, we use \texttt{gpt-4} \citep{openai2023gpt4} for both the baseline and LeanTutor. LeanTutor is given no student input and the system is asked to generate feeback types 2 (hint/question) and 3 (next step). We develop a simple baseline, providing the model with the erroneous student proof and prompting the model to generate the two feedback types. For the LeanTutor model, since there is no NL from the student in this case, we do not run the Autoformalizer or Next Step Generator. Instead, the system directly extracts the first step in the proof (in Lean) from the available \textit{staff-solution}, and this is passed to the Feedback Generation module. We evaluate 18 cold start proofs, two from each world. The results (Table \ref{coldstart}) indicate that LeanTutor outperforms the baseline on Accuracy and Relevance axes. 

\begin{table*}
    \centering
\begin{tabular}{lllll}
     \toprule
        \multicolumn{1}{c}{\bf Feedback Type} & \multicolumn{1}{c}{\bf Accuracy} & \multicolumn{1}{c}{\bf Relevance} & \multicolumn{1}{c}{\bf Readability} & \multicolumn{1}{c}{\bf Answer Leakage} \\
        \midrule
        Baseline Hint/Question & 3.6 & 3.2 & \textbf{4.9} & \textbf{4.5} \\
        LeanTutor Hint/Question & \textbf{4.3} & \textbf{4.4} & 4.4 & 4.1 \\
        \midrule
        Baseline Next Step & 3.6 & 3.2 & 4.9 & N/A \\
        LeanTutor Next Step & \textbf{3.9} & \textbf{4.8} & 4.9 & N/A \\
        
        \bottomrule
    \end{tabular}
    \caption{Average (across all proofs) qualitative scores of generated feedback from baseline and LeanTutor experiments on 18 cold-start proofs. A score closer to 5 indicates desired performance.}
    \label{coldstart}
\end{table*}

\section{LLM-as-a-Judge Evaluations}
Table \ref{tab:LLM-eval} contains the \texttt{gpt-4} ratings of the generated feedback from LeanTutor and the baseline models. The baseline model is scored more highly on nearly every axis and every feedback type. However, given that these results are quite different (even the general trends don't follow) from our evaluators' scores, we do not regard them in our final evaluation. This experiment provides some insight into the efficacy of LLM-as-a-judge evaluations of tutor hint generations. 

\begin{table*}[]
    
    \centering
    \begin{tabular}{lllll}
     \toprule
        \multicolumn{1}{c}{\bf Feedback Type} & \multicolumn{1}{c}{\bf Accuracy} & \multicolumn{1}{c}{\bf Relevance} & \multicolumn{1}{c}{\bf Readability} & \multicolumn{1}{c}{\bf Answer Leakage} \\
        \midrule
        Baseline Error Identification & \textbf{4.4} & 4.7 & \textbf{5} & \textbf{5} \\
        LeanTutor Error Identification & \textbf{4.1} & \textbf{4.3} & 4.9 & 5 \\
        \midrule
        Baseline Hint/Question & 4.7 & 4.7 & \textbf{5} & 4.9 \\
        LeanTutor Hint/Question & \textbf{4.4} & \textbf{4.4} & \textbf{4.8} & \textbf{4.9} \\
        \midrule
        Baseline Next Step & 4.3 & 4.5 & \textbf{4.8} & \textbf{3.0} \\
        LeanTutor Next Step & \textbf{3.9} & \textbf{4.0} & 4.6 & 2.6 \\
        
        \bottomrule
    \end{tabular}
    \caption{Average (across all proofs) scores of generated feedback from baseline and LeanTutor experiments on 21 incorrect proofs. The generated feedback was scored by an \texttt{gpt-4} on a scale of 1-5 in which a score closer to 5 indicates desired performance.}
    \label{tab:LLM-eval}
    
\end{table*}

\newpage
\newpage
\section{Model Prompts}
\subsection{Autoformalizer Prompts}
\label{appendix:autoformprompt}
\textit{Autoformalizer prompt for step-by-step formalization} contains the system and user prompts used for autoformalization. The following were given as input to the system prompt: the theorem statement of the proof (in NL and FL), the theorem and tactic dictionaries, five hard-coded examples of the autoformalization task, and the staff solution. The prompt for full proof generation, in \textit{Autoformalizer prompt for whole proof formalization},  is the same, except the five hard-coded examples were adjusted to whole proof translations, to match the whole proof autoformalization task. 

\begin{figure*}
\begin{tcolorbox}[
  colback=pink!20,
  colframe=pink!80,
  coltitle=black,
  fonttitle=\bfseries,
  title=Autoformalizer prompt for step-by-step formalization,
  fontupper=\ttfamily\normalsize,
  sharp corners,
]

\texttt{\#\#\# System:} \\[2pt]
\texttt{An undergraduate student is proving the following Peano Arithmetic theorem:} \\ 
\texttt{Theorem statement in natural language: \{theorem\_statement\_NL\}} \\ 
\texttt{Theorem statement in formal language: \{theorem\_statement\_FL\}} \\
Convert the student’s natural language mathematical proof step to Lean4 syntax. \\
\textit{[If \texttt{staff\_solution} is provided]} \\ 
\texttt{This is one example of the completed proof in Lean4, with in-line comments of the natural language proof corresponding to the Lean4 syntax: 
} \\ whole\_theorems[theorem\_name] \\

\texttt{These are the formal theorems you have access to:}  \\ 
\{theorem\_dict\} \\

These are the Lean tactics you have access to:  \\ 
\{tactic\_dict\} \\

Your response must be written as a single line of Lean tactic code, as used in the body of a by block of a Lean theorem.It should match the structure of Lean DSL tactic proofs, such as: \\
intro h  \\
rw [← is\_zero\_succ a] \\
apply succ\_inj at h \\
exact h \\
contrapose! h \\

Note: Only 1 lean tactic, do not write multiple lean tactics that are comma seperated. \\ 
DO *NOT* wrap your answer in markdown syntax, e.g. '''lean '''. It must be simply a Lean tactic script that can be inserted into a proof. \\
  
Here are some examples. NOTE: These are just examples. The correct Lean4 code may not necessarily use the propositions shown in these proofs. \\

Example 1: \\
Input: Rewrite the LHS pred (succ a) with the given statement that succ a = succ b, LHS is now pred (succ b) \\
Output: rw [h] \\ 

Example 2: \\
Input: Rewrite LHS using the commutative property of addition, changing a + (b + c) to a + b + c \\
Output: rw [← add\_assoc] \\ 

Example 3:  \\
Input: Assume that the hypothesis 'h' is true, that is, a + succ d = 0. The goal now is to prove that a = 0. \\
Output: rw [add\_zero] at h \\

Example 4: \\
Input: Split the natural number 'b' into two cases: 'b' is zero, and 'b' is the successor of another natural number 'd'. \\
Output: cases b with d \\

Example 5: \\
Input: Use the case of a + b to simplify the goal to equal z = x + (a + b). \\ 
Output: use a + b
 \\

\#\#\# User: The natural-language statement to formalize is: \\ 
\{nl\_statement\}

\end{tcolorbox}
\end{figure*}

All strings in typewriter font are runtime placeholders.  
\texttt{\{theorem\_statement\_NL\}} – theorem in natural language;  
\texttt{\{theorem\_statement\_FL\}} – the same theorem in Lean’s formal syntax;   
\texttt{\{whole\_theorems[theorem\_name]\}} – The staff solution;  
\texttt{\{theorem\_dict\}} – dictionary of Peano-arithmetic facts available to the model;  
\texttt{\{tactic\_dict\}} – dictionary of Lean tactics the model may use;  
\texttt{\{prev\_goal\}} – current Lean proof state ;  
\texttt{\{prev\_nl\}} – previous student proof lines ;  
\texttt{\{nl\_statement\}} – the natural-language step to be converted.  
The optional block, corresponding to the staff solution, renders optionally.

\begin{figure*}
\begin{tcolorbox}[
  colback=pink!20,
  colframe=pink!80,
  coltitle=black,
  fonttitle=\bfseries,
  title=Autoformalizer prompt for whole proof formalization,
  fontupper=\ttfamily\normalsize,
  sharp corners,
]

\texttt{\#\#\# System:} \\[2pt]
\texttt{An undergraduate student is proving the following Peano Arithmetic theorem:} \\ 
\texttt{Theorem statement in natural language: \{theorem\_statement\_NL\}} \\ 
\texttt{Theorem statement in formal language: \{theorem\_statement\_FL\}} \\[6pt]

Convert the student’s natural language mathematical proof to Lean4 syntax. \\[8pt]

\textit{[If \texttt{staff\_solution} is provided]} \\ 
\texttt{This is one example of the completed proof in Lean4, with in-line comments of the natural language proof corresponding to the Lean4 syntax: 
} \\ whole\_theorems[theorem\_name] \\[8pt]

\texttt{These are the formal theorems you have access to:}  \\ 
\{theorem\_dict\} \\[6pt]

These are the Lean tactics you have access to:  \\ 
\{tactic\_dict\} \\[8pt]

Your response must be written as a proof in Lean, in a list of tactics on each new line. SUch as: \\
intro h  \\
rw [← is\_zero\_succ a] \\
apply succ\_inj at h \\
exact h \\
contrapose! h \\ \\

Each tactic must be formatted consistently with Lean4's syntax and DO NOT include any comments in the list. \\ 
DO *NOT* wrap your answer in markdown syntax, e.g. '''lean '''. It must be simply a list of Lean tactics separated by \textbackslash n.
\\ \\
Here are some examples. NOTE: These are just examples. The correct Lean4 code may not necessarily use the propositions shown in these proofs. \\ \\

Example 1: \\
Input: Induct on b, with d = 0 as the base case and the inductive hypothesis a * d = d * a. There are now two proof goals, prove base case: a * 0 = 0 * a, and inductive step: a * succ d = succ d * a. \\
First we prove base case. \\
Simplify RHS 0 * a to 0. \\
Simplify LHS a * 0 to 0. \\
Prove LHS and RHS are equal, 0 = 0, completing base case. \\
Next prove inductive step. Rewrite RHS succ d * a to d * a + a. \\
Rewrite the RHS from d * a + a to a * d + a using the inductive hypothesis. \\
Rewrite the LHS, changing a * succ d to a * d + a. \\
Prove LHS and RHS are equal, a * d + a = a * d + a, completing the proof. \\
Output: induction b with d hd \\
rw [zero\_mul] \\
rw [mul\_zero] \\
rfl \\
rw [succ\_mul] \\
rw [← hd] \\ 
rw [mul\_succ] \\
rfl
\\[8pt]

\end{tcolorbox}

\end{figure*}

\begin{figure*}
\begin{tcolorbox}[
  colback=pink!20,
  colframe=pink!80,
  coltitle=black,
  fonttitle=\bfseries,
  title=Autoformalizer prompt for whole proof formalization (continued),
  fontupper=\ttfamily\normalsize,
  sharp corners,
]

Example 2: \\
Input: We must assume succ (succ 0) + succ (succ 0) = succ (succ (succ (succ (succ 0)))) and derive a contradiction or falsehood. \\
Using our previous theorems, we can change succ (succ 0) + succ (succ 0) into succ (succ (succ (succ 0))). \\
By the injectivity of succ, we know that 0 = succ 0. 
0 is not equal to the successor of any natural number, so we have a contradiction. \\
Thus, we have a falsehood/contradiction, which is what we wanted to show. \\
Output: intro h \\
rw [add\_succ, add\_succ, add\_zero] at h \\
repeat apply succ\_inj at h \\
apply zero\_ne\_succ at h \\
exact h
\\

Example 3: \\
Input: We consider the case where the successor of x is less than or equal to the successor of y. This implies that the successor of y is equal to the successor of x plus some natural number d. \\
We assume d as the difference such that when added to x results in y. The goal now is to prove that y is equal to x plus d. \\
We rewrite the right-hand side of succ y = succ x + d using the theorem that states the the successor of a sum of two natural numbers is the same as the successor of the first number added to the second number. \\
We apply the property that if two natural numbers with successors are equal, then the original numbers are also equal. \\
We have shown that x = y + d, so we can use this to prove the goal. \\
Output: cases hx with d hd \\
use d \\
rw [succ\_add] at hd \\
apply succ\_inj at hd \\
exact hd
\\

Example 4:
Input: We use proof by contraposition. So, we assume succ m = succ n and show m = n. \\
By the injectivity of succ, we have m = n. \\
So, m = n, which is exactly what we wanted to show. \\
Output: contrapose! h \\
apply succ\_inj at h \\
exact h
\\

\end{tcolorbox}

\end{figure*}

\begin{figure*}
\begin{tcolorbox}[
  colback=pink!20,
  colframe=pink!80,
  coltitle=black,
  fonttitle=\bfseries,
  title=Autoformalizer prompt for whole proof formalization (continued),
  fontupper=\ttfamily\normalsize,
  sharp corners,
]
Example 5: \\
Input: Rewrite the expression for the square of (a + b), a\^ 2, and b\^2 to be (a + b) * (a + b), a * a, and b * b respectively. \\
Rearrange the terms on the right hand side of the equation, swapping the order of b * b and 2 * a * b. This is based on the commutative property of addition, which states that the order of the terms does not change the result of the addition. \\
Rewrite the left-hand side of the equation using the distributive property of multiplication over addition. This expands (a + b) * (a + b) to a * a + b * a + a * b + b * b. \\
Rewrite the term 2 * a * b in the goal as (a * b + a * b) using the theorem that 2 times a number is the same as the number added to itself. Also, rewrite the term a * b + b * b as (a * b + a * b) + b * b using the theorem that the product of a sum is the sum of the products. \\
We rewrite the expression a * b as b * a in the goal. This is based on the commutative property of multiplication, which states that the order of the factors does not change the product. This results in the new goal: a * a + a * b + (a * b + b * b) = a * a + (a * b + a * b) + b * b. \\
We use the theorem that states the associativity of addition twice to rearrange the left-hand side of the equation. This changes the goal to proving that a * a + a * b + a * b + b * b equals a * a + a * b + a * b + b * b. \\
The goal is now to prove that a * a + a * b + a * b + b * b = a * a + a * b + a * b + b * b, which is true by reflexivity \\
Output: rw [pow\_two, pow\_two, pow\_two] \\
rw [add\_right\_comm] \\ 
rw [mul\_add, add\_mul, add\_mul] \\
rw [two\_mul, add\_mul] \\
rw [mul\_comm b a] \\
rw [← add\_assoc, ← add\_assoc] \\
rfl

\#\#\# User: The natural language proof that we want to formalize: \\ 
\{nl\_statement\}

\end{tcolorbox}
\caption{All strings in typewriter font are runtime placeholders.  
\texttt{\{theorem\_statement\_NL\}} – theorem in natural language;  
\texttt{\{theorem\_statement\_FL\}} – the same theorem in Lean’s formal syntax;   
\texttt{\{whole\_theorems[theorem\_name]\}} – The staff solution;  
\texttt{\{theorem\_dict\}} – dictionary of Peano-arithmetic facts available to the model;  
\texttt{\{tactic\_dict\}} – dictionary of Lean tactics the model may use;  
\texttt{\{prev\_goal\}} – current Lean proof state ;  
\texttt{\{prev\_nl\}} – previous student proof lines ;  
\texttt{\{nl\_statement\}} – the natural-language proof to be converted.  
The optional block, corresponding to the staff solution, renders optionally. }

\end{figure*}

\label{fig:autoformalizerpromptfull}

\vspace{1em}

\subsection{Natural Language Feedback Generation}
\label{appendix:feedbackgenprompt}
\textit{Natural language feedback generation prompt} is the prompt to generate student feedback for incorrect proof inputs. This prompt is used in our final end-to-end system evaluation. 
\begin{figure*}
\begin{tcolorbox}[colback=yellow!20, colframe=yellow!50, coltitle=black, fonttitle=\bfseries,   title= Natural language feedback generation prompt]

\texttt{\#\#\# System}: \texttt{You are a  math professor, identifying the error in student proofs, with the help of the Lean4 verifier.} \\ \\
\texttt{\#\#\# User}: \texttt{A first-year math student's incomplete Peano Arithmetic proof has been formalized in Lean4, but it has an error.} \\
    \texttt{This is the incorrect student proof in Lean4: } \\ \\
    \texttt{\{lean\_proof\}} \\ 

    \texttt{This is the current Lean4 state, throwing an error due to the last step {last\_line}:} \\ 
    
    \texttt{\{error\}} \\

    \texttt{The actual correct step in Lean4 is:}  \\

    \texttt{\{next\_step\}} \\

    \texttt{Error Categories include:} \\
    \texttt{1. Inducting on the incorrect variable} \\
    \texttt{2. Selecting the incorrect base case } \\
    \texttt{3. Not generalizing the inductive step to all cases }\\
    \texttt{4. Failing to apply the inductive hypothesis }\\
    \texttt{5. Incorrect/Incomplete simplification or expansion }\\
    \texttt{6. Incorrect calculation or careless mistake}\\
    \texttt{7. Other} \\

    \texttt{Explain the student error, ask a guiding question to reach correct next step, and give a hint that explicitly reveals the answer in 1-2 sentences. Be specific and use equations from goal states. } \\ 
    
    \texttt{DO NOT USE any "Lean" or any Lean tactics or syntax such as "tactic" or "reflexivity" or theorems such as "add\_comm". You are speaking directly to the student, use "You" language. } \\

    \texttt{Example:} \\

    \texttt{Type: Incorrect simplification} \\
    \texttt{Message: The RHS of your equation, a + (b + succ d), cannot be simplified with your applied strategy.}
    \texttt{Question/Hint: Do you know of a theorem that can perform this simplification?
    Informalization: The next step is to rewrite a + (b + succ d) as (a + b) + succ d.} \\

    \texttt{IMPORTANT: Respond with ONLY a raw JSON object in the following format, without any code block formatting or additional text:} \\
    \texttt{\{}\\
    \texttt{"Type": "Students' error type",}\\
    \texttt{"Message": "Brief description of error in this problem"}\\
    \texttt{"Question": "Do you....?"}\\
    \texttt{"Informalization": "The next step is to..."}\\
    \texttt{\}}
\end{tcolorbox} 
\caption{\texttt{\{lean\_proof\}} is a placeholder for the autoformalized proof until now. \texttt{\{error\}} is the Lean compiler error thrown by the formalized proof. \texttt\{next\_step\} is a placeholder for the next tactic generated by the NSG module.}
\end{figure*}

\label{fig:inform_incorrect_prompt}

\subsection{Baseline Prompt for Full System Evaluation}
\label{appendix:baselinefulleval}
\textit{Natural language error + next-step prompt} is the baseline prompt used in end-to-end system evaluation. This prompt does not receive any Lean inputs. 
\begin{figure*}
\begin{tcolorbox}[colback=orange!20, colframe=orange!50, coltitle=black, fonttitle=\bfseries, title= Natural language error + next-step prompt]

\texttt{\#\#\# System}: \texttt{You are a math professor helping a student debug their Peano Arithmetic proof.} \\\\

\texttt{\#\#\# User}: \texttt{A first-year math student is working on the following Peano Arithmetic theorem:} \\
\texttt{\{theorem\}} \\\\

\texttt{Below are the steps of the proof the student has completed thus far. There may be errors and/or the work may be incomplete:} \\
\texttt{\{proof\}} \\\\

\texttt{Identify and explain the student error, if it exists. Then, identify the correct next step. Ask a guiding question or give a hint that can help the student reach the correct next step in 1-2 sentences. Be specific.} \\\\

\texttt{Speak directly to the student using "You" language. Avoid using Lean tactics or syntax like "apply", "intro", or "rw".} \\\\

\texttt{Example:} \\
\texttt{Error Message: The RHS of your equation, a + (b + succ d), cannot be simplified with your applied strategy.} \\
\texttt{Next Step: The next step is to rewrite a + (b + succ d) as (a + b) + succ d.} \\
\texttt{Question/Hint: Do you know of a theorem that can perform this simplification?} \\\\

\texttt{IMPORTANT: Respond with ONLY a raw JSON object in the following format, without any code block formatting or additional text:} \\
\texttt{\{} \\
\texttt{"Error\_Message": "Brief description of error in this problem",} \\
\texttt{"Next\_Step": "The next step is to..."}, \\
\texttt{"Question": "Do you....?"} \\
\texttt{\}}

\end{tcolorbox}
\caption{\texttt{\{theorem\}} is a runtime placeholder for the theorem statement (in NL). \texttt{\{proof\}} is a placeholder for the student’s current attempt.}
\end{figure*}

\label{fig:nl_error_next_step_prompt}

\bibliography{main}


\end{document}